\documentclass[acmsmall]{acmart}
\usepackage{bbm}
\usepackage{arabtex}
\usepackage{utf8}

\AtBeginDocument{%
  \providecommand\BibTeX{{%
    \normalfont B\kern-0.5em{\scshape i\kern-0.25em b}\kern-0.8em\TeX}}}

\setcopyright{acmlicensed}
\copyrightyear{2018}
\acmYear{2018}
\acmDOI{XXXXXXX.XXXXXXX}

\acmJournal{TALLIP}
\acmVolume{37}
\acmNumber{4}
\acmArticle{111}
\acmMonth{8}

\begin{document}
\vocalize
\setarab
\title{AraSpot: Arabic Spoken Command Spotting}

\author{Mahmoud Salhab}
\email{Mahmoud.Salhab@lau.edu.lb}
\orcid{0009-0005-8309-9230}
\author{Haidar Harmanani}
\orcid{0000-0001-5416-4383}
\email{haidar@lau.edu.lb}
\affiliation{%
  \institution{Lebanese American University}
  \streetaddress{Department of Computer Science and Mathematics}
  \city{Byblos}
  \country{Lebanon}
  \postcode{1401 2010}
}

\renewcommand{\shortauthors}{Salhab and Harmanani}

\begin{abstract}
Spoken keyword spotting (KWS) is the task of identifying a keyword in an audio stream and is widely used in smart devices at the edge in order to activate voice assistants and perform hands-free tasks.  The task is daunting as there is a need, on the one hand, to achieve high accuracy while at the same time ensuring that such systems continue to run efficiently on low power and possibly limited computational capabilities devices.  This work presents AraSpot for Arabic keyword spotting trained on 40 Arabic keywords, using different online data augmentation, and introducing ConformerGRU model architecture.  Finally, we further improve the performance of the model by training a text-to-speech model for synthetic data generation. AraSpot achieved a State-of-the-Art SOTA 99.59\% result outperforming previous approaches.\footnote{Available on GitHub at \url{https://github.com/msalhab96/AraSpot}}
\end{abstract}



\received{20 February 2007}
\received[revised]{12 March 2009}
\received[accepted]{5 June 2009}

\maketitle

\section{Introduction}

Automatic Speech Recognition (ASR) is a fast-growing technology that has been attracting increased interest due to its embedment in a myriad of devices.  ASR allows users to activate voice assistants and perform hands-free tasks by detecting a stream of input speech and converting it into its corresponding text. Spoken Keyword Spotting (KWS) is similar to the ASR problem but it is mostly concerned with the identification of predefined keywords in continuous speech \cite{6854211}. In fact, keyword spotting systems are common components in speech-enabled devices \cite{8268946} and have a wide range of applications such as speech data mining, audio indexing, phone call routing, and many other \cite{inproceedings}.

Recently, many models became popular for tackling KWS systems including \textit{Convolution Neural Networks} (CNN), \textit{Residual Networks} (ResNet), and \textit{Recurrent Neural Networks} (RNN).  The disadvantage of CNN is that they do not work well with sequences.  Furthermore, CNNs are not usually able to capture long term-dependencies in human speech signal, the same thing for ResNets as they are short-sighted when it comes to their respective field.  On the other hand, recurrent neural network directly models the input sequence without learning local structure between successive time series and frequency steps \cite{8607038}.

The Google Speech Command (GSC) datasets \cite{gsc} is the de facto KWS standard for English. Unfortunately, KWS has considerably lesser publicly available data than ASR, Consequently, training a neural network becomes harder given the scarcity of the data available \cite{9427206}. To overcome data scarcity for KWS, many researchers are using pre-trained models and synthesized data such as in \cite{https://doi.org/10.48550/arxiv.2002.01322}.

Most KWS research has focused on English and Asian languages with few research investigating KWS in Arabic, despite the fact that Arabic is the $4\textsuperscript{th}$ mostly used language on the internet \cite{article, 6841973}.  This study introduces AraSpot for Arabic command spotting, leveraging the ASC dataset published in \cite{asc}. We explore various online data augmentation techniques to model diverse environmental conditions, thereby enhancing and expanding the dataset.  The proposed approach introduces a \textit{ConformerGRU} model architecture to address short and long dependency issues in both RNN and CNN. We demonstrate based on empirical evidence, that our proposed model architecture surpasses all previous approaches on the dataset. Furthermore, we enhance model performance by augmenting the training data with additional speakers through synthetic data generation. To our knowledge, this study is the first to implement the conformer architecture with a GRU layer for KWS on the ASC dataset, while also incorporating synthetic data generation techniques.

This paper is organized as follows. Section \ref{sec:related} presents a literature review, followed by our methodology in section \ref{sec:methodology}. Section \ref{sec:results} presents the experiments and the results, and lastly, we conclude in section \ref{sec:conc} with a summary of potential future work.

\section{Related Work}
\label{sec:related}
Keyword spotting received a considerable amount of interest from the research community.  One of the earliest approaches is based on the use of large-vocabulary continuous speech recognition (LVCSR).  In such systems, the speech signal is first decoded and then searched for the keyword/filler in the generated lattices \cite{319341, 6707766, 10.21437}.  An alternative to LVCSR is the keyword Hidden Markov Model (HMM) where a keyword HMM and a filler HMM are trained to model keyword and non-keyword audio segments \cite{115555, 266505}.

With the rise of GPU computational power and the increase in data availability, the research community switched gears towards deep learning-based KWS systems.  For example, Coucke et al. \cite{https://doi.org/10.48550/arxiv.1811.07684} used dilated convolutions of the WaveNet architecture and showed that the results were more robust in the presence of noise than LSTM or CNN based models. Arik et al. \cite{https://doi.org/10.48550/arxiv.1703.05390} proposed a single-layer CNN and two-layer RNNs, similarly, two gated CNN with one-layer bi-directional LSTM proposed in \cite{GCNNLSTM}. An attention-based end-to-end model introduced for small-footprint KWS was proposed in \cite{https://doi.org/10.48550/arxiv.1803.10916}. To overcome KWS data scarcity, Sun et al. \cite{Sun2017} used transfer learning  by  training an ASR system, and the acoustic model of the trained ASR model was fine-tuned on the KWS task.

Lin et al. \cite{https://doi.org/10.48550/arxiv.2002.01322} showed that building a state-of-the-art (SOTA) KWS model requires more than 4000 utterances per command.  The authors also noted that with the various limitations and difficulties in acquiring more data, methods to enlarge and expand the training data are required.  The above problem was alleviated in \cite{https://doi.org/10.48550/arxiv.1909.11699, https://doi.org/10.48550/arxiv.1811.00707} by using synthesized speech through data augmentation approaches.  The method utilized a text-to-speech system in order to generate synthetic speech. Further more, to enhance the robustness of the model against different noisy environment, artificial data corruption by adding reverberated music
or TV/movie audio to each utterance at a certain speech-to-interference (SIR) ratio used in \cite{raju2018data}. Furthermore, the study conducted by researchers in \cite{7953152} explores the influence of data augmentation on speech recognition system performance. This is achieved through the generation of far-field data using simulated and real room impulse responses (RIR), specifically utilizing reverberation techniques. Moreover, a room simulator developed in  \cite{46107} is used to generate large-scale simulated data for training deep neural networks for far-field speech recognition, this simulation-based approach was employed in Google Home product and brought significant performance improvement. 

For Arabic, Ghandoura et al. \cite{asc} recorded and published a benchmark that includes 40 commands that were recorded by 30 different speakers.  The authors achieved 97.97\% accuracy using a deep CNN model, and to increase the data diversity the researcher used different data augmentation techniques. Benamer et al. \cite{asc2} published another benchmark that included 16 commands but used an LSTM model instead. Furthermore, a keyword spotting system was presented in \cite{arabicaudsearch} to perform audio searching of uttered words in Arabic speech.

\section{Dataset Description}

The ASC dataset \cite{asc} includes $12,000$ pairs of one-second-long audio files and corresponding keywords, totaling $40$ keywords. Each keyword has $300$ audio files recorded by $30$ participants, each providing $10$ utterances per keyword.  Some of the keywords in the ASC dataset were inspired by the Google Speech Commands (GSC) \cite{gsc} dataset, while the remaining commands were selected to be grouped into broad and potentially overlapping categories.  The dataset has $300$ utterances per keyword for a total size of $384$ MB.  Criteria for audio file settings include a sampling rate of 16 kHz, 16 bits per sample, mono-signal, and a \textit{.wav} file format. The dataset is in standard Arabic, with all recordings were done using a laptop with an external microphone in a quiet environment. The keywords have been chosen to activate voice assistants and perform hands-free task for some applications and devices such as a simple photo browser or a keypad \cite{asc}.  Table~\ref{commands} lists the 40 keywords in the dataset and their Arabic translation.  It should be noted that the ASC dataset exhibits fewer utterances per class but cleaner data quality due to manual segmentation.

\begin{table}[!h]
\setcode{utf8}
  \begin{center}
      \centering
      \begin{tabular}{c|c}
      \end{tabular}
    \begin{tabular}{c|c||c|c}
      \hline
      Translation & Keyword & Translation & Keyword \\ \hline \hline
      Zero &  \RL{صفر} & Enable  & \RL{تفعيل}\\ \hline
      One & \RL{واحد}  & Disable  & \RL{تعطيل}\\ \hline
      Two & \RL{اثنان} & Ok  & \RL{موافق}\\ \hline
      Three & \RL{ثلاثة}  & Cancel  & \RL{إلغاء}\\ \hline
      Four & \RL{أربعة}  & Open  & \RL{فتح}\\ \hline
      Five & \RL{خمسة}  & Close  & \RL{إغلاق}\\ \hline
      Six & \RL{ستة} & Zoom in  & \RL{تكبير}\\ \hline
      Seven & \RL{سبعة} & Zoom Out  & \RL{تصغير}\\ \hline
      Eight & \RL{ثمانية} & Previous  & \RL{السابق}\\ \hline
      Nine & \RL{تسعة} & Next  & \RL{التالي}\\ \hline
      Right & \RL{يمين} & Send  & \RL{إرسال}\\ \hline
      Left & \RL{يسار} & Receive  & \RL{استقبال}\\ \hline
      Up & \RL{أعلى} & Move  & \RL{تحريك}\\ \hline
      Down & \RL{أسفل} & Rotate  & \RL{تدوير}\\ \hline
      Front & \RL{أمام} & Record  & \RL{تسجيل}\\ \hline
      Back & \RL{خلف} & Enter  & \RL{إدخال}\\ \hline
      Yes & \RL{نعم} & Digit  & \RL{رقم}\\ \hline
      No & \RL{لا} & Direction  & \RL{اتجاه}\\ \hline
      Start & \RL{ابدأ} & Options  & \RL{خيارات}\\ \hline
      Stop & \RL{توقف} & Undo  & \RL{تراجع}\\ \hline
    \end{tabular}
    \caption{\label{commands} The 40 commands used in the ASC dataset}
    \label{dataset}
  \end{center}
  \end{table}

\section{Solution Approach}

\label{sec:methodology}

\subsection{Data Augmentation}
\label{data-aug}

The core idea of data augmentation is to generate additional synthetic data to improve the data diversity to cover comprehensive range of conditions that could potentially be present in any unseen instance.  The augmented data is typically viewed as belonging to a distribution that is close to the original one \cite{dataaug}, while the resulting augmented examples can be still semantically described by the labels of the original input examples which is known as label-preserving transformation.  Augmented data is normally generated on the fly during the training process in what is known as \textit{online augmentation}.  Another alternative is  \textit{offline augmentation} \cite{dataaug1} which transforms the data beforehand and stores it in memory.

For this work, we apply on-the-fly data augmentation, in both the time domain as well as the frequency domain. Let $F_t$ and $F_f$ be a set of pre-defined time domain and frequency domain transformation/augmentation functions such that $F_t=\{f_1, f_2, \ldots,f_Q\}$, and $F_f=\{f_1, f_2, \ldots,f_V\}$, for a given input speech signal $x_i$ we first apply the chosen time-domain augmentation $\tilde{F_{t}^{i}}$ for the $i^{th}$ signal, then after transforming the augmented signal into the frequency domain, we apply the chosen frequency augmentation $\tilde{F_{f}^{i}}$:

\begin{equation}
    \tilde{F_{t}^{i}} = \{f_q: r_{q}^{i} \geq \lambda, 1 \leq q \leq Q\}
    \label{timeaug}
\end{equation}

\begin{equation}
    \tilde{F_{f}^{i}} = \{f_v: r_{v}^{i} \geq \gamma, 1 \leq v \leq V\}
    \label{freqaug}
\end{equation}

Where $r_{v}^{i}$ and $r_{q}^{i}$ represent uniformly sampled values from $[0, 1]$ at each training step for each augmentation operation, and $\tilde{F}{t}^i$ and $\tilde{F}{f}^i$ denote the time domain and frequency domain functions with operation order shuffling per domain applied on the $i^{th}$ input signal at a given training step. Finally, $\lambda$ and $\gamma$ denote the time domain and frequency domain augmentation rates, ensuring that any signal can have any possible augmentation combination with different orders from one epoch to the next.

For a given speech signal $X$ in the time domain, the below time domain augmentation methods are used as items for $F_t$:

\begin{enumerate}
  \item \textit{Urban Background Noise Injection}:  We used noise injection similar to \cite{asc}, but we used the test set of the \textit{Freesound data} published in \cite{inproceedings}. We first concatenated all existing $K$ noise audios into a single noise signal $\mathcal{N}$, and apply the the augmentation process as follows:

\begin{equation}
    m \sim unif(0, T_{n})
  \label{bgnoise1}
\end{equation}

\begin{equation}
n \sim unif(m, min(T_{n}, m + T_{s}))
  \label{bgnoise2}
\end{equation}

\begin{equation}
f \sim unif(0, T_s - n + m - 1)
  \label{bgnoise2}
\end{equation}

\begin{equation}
\xi = [0]_{f} \parallel (\mathcal{N}_{i})_{m \leq i < n}  \parallel [0]_{T_s - f - n + m }
  \label{bgnoise3}
\end{equation}

\begin{equation}
\acute{X}=\mathcal{G} \xi + X
  \label{bgnoise3}
\end{equation}

Where $T_{s} = \mid X\mid$, $T_{n} = \mid \mathcal{N}\mid$, $n$ and $m$ represent the start and end of the noise segment in $\mathcal{N}$, while $f$ denotes the degree of freedom ensuring variability in the starting point of addition for the same audio across different steps. Additionally, $\xi$ denotes the noise segment, $\parallel$ signifies the concatenation operation, where the selected noise chunk is concatenated with leading and trailing zeros of size $f$, resulting in $T\textsubscript{s} - f - n + m$. It should be noted that $\acute{X}$ represents the augmented version of $X$, and $\mathcal{G}$ denotes a random gain between $0$ and $1$.

\item \textit{Speech Reverberation}: speech reverberation is originally caused by the environment surrounded by the source, where the end result received by the input device (i.e Microphone) is the sum of multiple shifted and attenuated signals of the same original signal \cite{s22020592}. To simulate speech reverberation, this can be done by convolving the original input speech signal with a room impulse response (RIR). For this case, we used both \textit{RIR} datasets created and published in \cite{rir1} and \cite{rir2}.

Let $H=\{h_1, h_2,\ldots,h_R\}$ be a set of all available impulse responses, where each one of 1-second length.  For a given speech signal $X$, the augmentation process is done as below: 

\begin{equation}
h \sim unif(H)
  \label{reverberation1}
\end{equation}

\begin{equation}
l \sim unif(a, b)
  \label{reverberation2}
\end{equation}

\begin{equation}
\acute{X}= X \ast (h_i)_{0 \leq i \leq l}
  \label{reverberation3}
\end{equation}

\begin{equation}
\acute{X}[n]= \sum_{i=0}^{l} h[i]X[n - i]
  \label{reverberation4}
\end{equation}

where $l$ is the speech reverberation length, $\ast$ symbol in Equation \ref{reverberation3} is the convolution operation, and $a$ and $b$ are the minimum and maximum reverberation length, we set $a$ to 31 ms, and $b$ to 250 ms.

\item \textit{Random Volume Gain:} Similar to the work done in \cite{asc}, for a given signal $X$, the magnitude of the signal is multiplied by a random gain $\mathcal{G}$ as shown below:

\begin{equation}
    \acute{X} = \mathcal{G} X
\end{equation}

where $\mathcal{G}$ is a random value between 0.2 and 2.

\item \textit{Random Fade In/Out:} Given a speech signal $X$ we multiply the magnitudes of the signal by a fade signal such as linear, exponential, logarithmic, quarter-sine, and half-sine.  The fade function is sampled uniformly from the previously mentioned signals, and then multiplied by the original signal.  The length of the fade signal is chosen randomly between 0 and $\mid X \mid$ and padded with ones to match the length of the original waveform $X$, and that can be formally shown as below:

\begin{equation}
    \acute{X} = F_{in} F_{out} X
\end{equation}

Where $F_{in}$ is the fade-in signal, $F_{out}$ is the fade-out signal.

\end{enumerate}

For a given signal $X$ in the frequency domain, spectrogram-based augmentation can be applied as proposed in \cite{SpecAugment}, for this work, we mainly used the time and frequency masking as items for $F_f$.

\subsection{Synthetic Data Generation Using TTS}
\label{synthaticdata}
End-to-end (E2E) Text-to-Speech (TTS) systems are used to generate speech directly from a given text, unlike traditional TTS systems that use complex pipelines. Seq2Seq-based TTS systems such as \cite{tts0, tts1, tts2, tts3} are commonly composed of an encoder, decoder, and an attention mechanism, such that, the characters embedding are projected into Mel-scale spectrogram followed by a vocoder that converts the predicted Mel-scale spectrogram into a waveform. 

In this work we use Tacotron 2 \cite{tts3} which has a relatively simple architecture.  The model consists of an \textit{encoder} and a \textit{decoder} with \textit{attention}.  The encoder takes the input characters/phonemes sequence $C$ and projects it into a high-level representation $h$, then the decoder with attention generates Mel-scale spectrogram frames by attending on $h$ and conditioning on the previously predicted frames.

We used the same setup used in the original Tacotron 2 paper \cite{tts3}.  Thus, we used WaveGlow \cite{waveglow} as a vocoder, and for the data to train the TTS on, we used the Arabic Common Voice dataset \footnote{https://voice.mozilla.org/}.  The data was filtered in order to use the top 10 speakers that have the highest number of utterances with relatively the highest quality.  This was done since most speakers in the dataset do not have a large number of utterances as training the model on small number of records per speaker leads to inconsistency, and the generated speech becomes unintelligible.

\subsection{ConformerGRU Model}

Convolution Neural network (CNN) and Recurrent Neural network (RNN) have their own advantages and limitations.  For example, while CNN exploits local information and local dependencies, RNN exploits long-term information and dependencies.

The Conformer architecture, as introduced in \cite{conformer}, has gained huge attention in various speech recognition applications, including those mentioned in \cite{10023291, Zhang_2023, 10.1007/978-3-031-20233-9_34}. This popularity is attributed to its unique capability, outlined in \cite{conformer}, to effectively capture information along with long and short-term dependencies. This is achieved through the fusion of multi-head self attention from the Transformer architecture \cite{transformer} with convolutional neural networks. Consequently, the resulting model is adept at modeling both local and global dependencies.

In the process of generating a latent vector representing the entirety of the input speech sequence, we employed a bidirectional Gated Recurrent Unit (GRU) layer. This configuration involves concatenating the latest hidden vectors from both the forward and backward directions, thereby treating the resulting concatenated vector as the latent representation of the input sequence, therefore, we combine the \textit{Conformer Block} with a \textit{Gated Recurrent Unit} GRU layer as described next. 

Given a dataset $\mathcal{D}=\{(x_1, y_1), (x_2, y_2),\ldots,(x_N, y_N)\}$ where $x_i$, and $y_i$ are the $i^{th}$ input example and the target label respectively, the objective is to model $P(Y \mid X)$ using a function $f_\theta$ that maximizes the following objective function:

\begin{equation}
\max_{\theta} \prod_{i=1}^{N} P(y_i \mid  x_i;\theta)
    \label{objective}
\end{equation}

\begin{equation}
\min_{\theta} \sum_{i=1}^{N} -log(P(y_i \mid  x_i;\theta))
    \label{entropy}
\end{equation}

\begin{figure}[t]
  \includegraphics[width=0.9\linewidth]{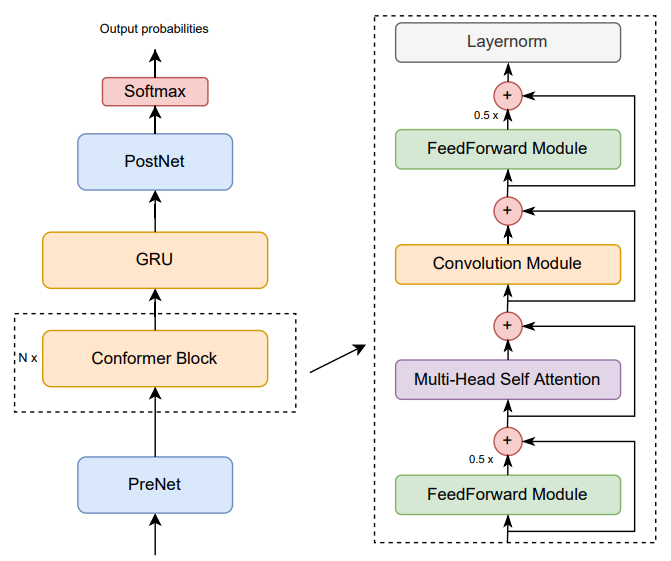}
  \caption{\label{model} ConformerGRU model architecture}
\end{figure}

To model $f_\theta$, we propose the \textit{ConformerGRU} model that consists of the following layers with the full architecture shown in Figure \ref{model}:
\begin{enumerate}
    \item A \textit{Pre-net Layer} that projects the speech feature space into a higher-level representation;
    \item A \textit{Conformer Block} that consists of multiple Conformer layers, where we can ensure the model able to handle long and short-term information dependencies.
    \item A single \textit{Gated Recurrent Unit (GRU)} which acts as an aggregate function instead of using the sum or the average of hidden states or the first hidden state only.
    \item A \textit{Post-net Layer} of two modules where the first is a simple projection layer followed by a prediction layer with a \textit{softmax} activation function.
\end{enumerate}

\section{Experiments and results}
\label{sec:results}
\subsection{Experiments Setup}

Let the data $\mathcal{D}=\{(x_{1}, y_{1}), (x_{2}, y_{2}),\ldots,(x_{N}, y_{N})\}$, such that $x_{i}$ and $y_{i}$ are the speech signal and the target label/command respectively.  Let $y_{i}\in{Y}$ and $x_{i}\in \mathbb{R}^{C \times S}$, where $Y$ is the set of all unique labels, $C$ is the number channels, and $S$ the number of the speech samples in that utterance.  We added an extra-label to represent the noise/NULL label.  Thus, 300 noise audios were generated and split into 60\% for training, 20\% for validation, and 20\% for testing, using the same noise audios and similar criteria as in \cite{asc}.

All the synthetic data generated from the text-to-speech mentioned in Section \ref{synthaticdata} for all speakers was added to the training data.  Furthermore, online augmentation as mentioned in Section \ref{data-aug} was applied during training and no offline augmentation was used.

For all experiments, we extracted 40 Mel-frequency cepstral coefficients (MFCC) features which were computed using a 25ms window size, with a stride of 10ms, and 80-channel filter banks.

We used the negative log-likelihood loss, and Adam optimizer with linear learning rate decay as shown in Equation \ref{lr} where $lr_0$ is the initial learning rate, for all experiments we set $lr_0$ to $10^{-3}$, $e \in [0, E)$ is the current epoch, and $E$ is the total number of epochs, and lastly, a dropout of 15\% ratio used for regularization purposes.

We trained all models on a single machine using a single NVIDIA 3080 TI GPU, with a batch size of 256.
Since the data is balanced across all labels, we used accuracy shown in Equation \ref{accuracy} as a metric to measure the performance across all experiments, given that $\hat{y}_i$ is the predicted class for the  $i^{th}$ example. 

\begin{equation}
    Accuracy = \frac{1}{N} \sum_{i=1}^N \mathbbm{1}(\hat{y}_i == y_i) * 100\%
\label{accuracy}
\end{equation}

\begin{equation}
    lr(e, E) = lr_0 * (1 - \frac{e}{E})
    \label{lr}
\end{equation}

\begin{table}[!htp]
  \begin{center}
      \centering
      \begin{tabular}{c|c}
      \end{tabular}
    \begin{tabular}{ccccc}
      \hline \hline
      $d_{model}$ & $h$ & $N$ & $ACC(\%)$ & $\#Params$\\ \hline \hline
      64 & 4 & 2 & 98.21 & 234K\\ \hline 
      64 & 4 & 1 & 97.19 & 165K\\ \hline 
      64 & 2 & 2 & $\boldsymbol{98.5}$ & 234K\\ \hline 
      64 & 2 & 1 & 97.64 & 165K\\ \hline \hline 
      96 & 4 & 2 & 98.78 & 511K\\ \hline
      96 & 4 & 1 & 98.17 & 358K\\ \hline 
      96 & 2 & 2 & $\boldsymbol{99.1}$ & 511K\\ \hline 
      96 & 2 & 1 & 98.17 & 358K\\ \hline \hline 
      128 & 4 & 2 & 99.17 & 895K\\ \hline 
      128 & 4 & 1 & 98.7 & 625K\\ \hline 
      128 & 2 & 2 & $\boldsymbol{99.35}$ & 895K\\ \hline 
      128 & 2 & 1 & 98.61 & 625K\\ \hline \hline 
    \end{tabular}
    \caption{\label{results} Results obtained by training the model on the original training data only, where $d_{model}$ is the model dimensionality, $h$ is the number of attention heads, $N$ is the number of conformer layers, $ACC$ is the accuracy, and lastly $\#Params$ is the number of model parameters.}
  \end{center}
  \end{table}

  \begin{table}[!htp]
  \begin{center}
      \centering
      \begin{tabular}{c|c}
      \end{tabular}
    \begin{tabular}{ccccc}
      \hline \hline
      d\_model & h & N & ACC(\%) & \#Params\\ \hline \hline
      64 & 4 & 2 & 98.41 & 234K\\ \hline 
      64 & 4 & 1 & 98.01 & 165K\\ \hline 
      64 & 2 & 2 & $\boldsymbol{98.66}$ & 234K\\ \hline 
      64 & 2 & 1 & 97.93 & 165K\\ \hline \hline 
      96 & 4 & 2 & $\boldsymbol{99.19}$ & 511K\\ \hline
      96 & 4 & 1 & 98.41 & 358K\\ \hline 
      96 & 2 & 2 & 99.15 & 511K\\ \hline 
      96 & 2 & 1 & 98.54 & 358K\\ \hline \hline 
      128 & 4 & 2 & 99.23  & 895K\\ \hline 
      128 & 4 & 1 & 98.94 & 625K\\ \hline 
      128 & 2 & 2 & $\boldsymbol{99.59}$ & 895K\\ \hline 
      128 & 2 & 1 & 99.27 & 625K\\ \hline \hline 
    \end{tabular}
    \caption{\label{synthresult} Results obtained by training the model on the original training data with the synthetic data combined, where $d_{model}$ is the model dimensionality, $h$ is the number of attention heads, $N$ is the number of conformer layers, $ACC$ is the accuracy, and lastly $\#Params$ is the number of model parameters.}
  \end{center}
  \end{table}

  \begin{figure}[!htp]
  \includegraphics[width=\linewidth]{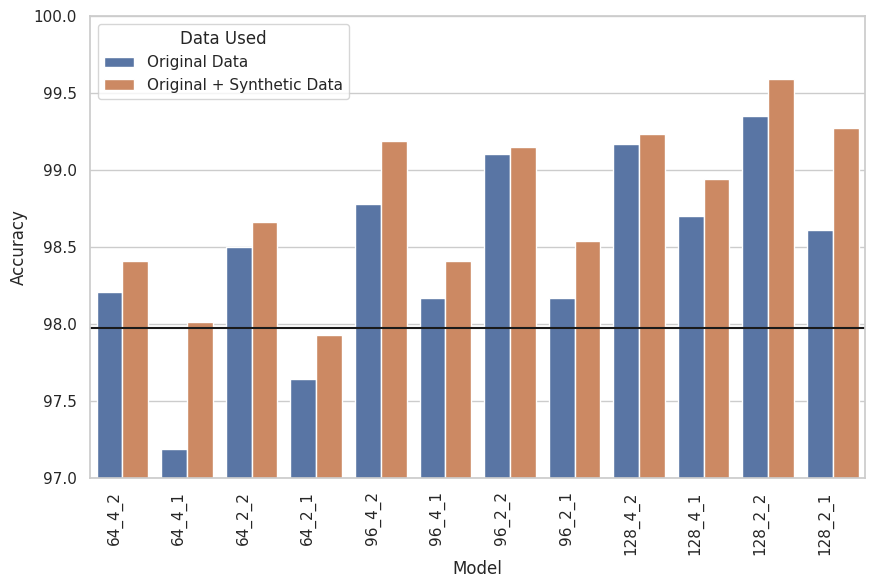}
  \caption{\label{graph:results} Analysis of AraSpot performance under various scenarios, illustrating the model parameters (dimensionality, number of heads, and layers) on the X-axis and corresponding accuracy on the Y-axis. The horizontal black line represents the accuracy of the optimal model from the literature \cite{asc}. Results are presented for models trained on original data with synthetic data generated through TTS and online data augmentation (depicted by blue bars), as well as models trained solely on original data with online data augmentation (depicted by orange bars).}
\end{figure}

\subsection{Results}

Multiple experiments were conducted to assess the impact of different parameters on the accuracy. This involved exploring variations in attention heads, model dimensionality, and conformer layers. Specifically, our investigation focused on examining how changes in the number of conformer layers, self-attention heads, and model dimensionality affect the system's performance.

We examine the performance change while only using data augmentation as detailed in Section \ref{data-aug} and without using additional synthetic data generation.  As shown in Table \ref{results}, increasing the model's dimensionality was found to enhance the performance, while a higher number of attention heads did not yield improved results. The additional attention heads did not lead to further improvements in the results because they failed to provide new or useful information beyond what was already captured by the existing attention mechanisms. This redundancy in information contributed to the diminishing returns observed in performance improvement.

In addition to that, for any given model dimensionality $d_{model}$ and number of self attention heads $h$, it is always the case that having higher number of conformer layer (i.e having $N=2$) gives higher accuracy. 

The introduction of synthetic data through TTS significantly enhanced the model performance in all scenarios, as evident in Table \ref{synthresult} and Figure \ref{graph:results}. 

In terms of model architecture, the (128, 2, 2) configuration for ($d_{model}$, h, N) consistently yields optimal results, whether synthetic data is employed or not. In Table \ref{results}, across all $d_{model}$ values, the best (h,N) combination is always (2,2). In Table \ref{synthresult}, this combination also shows high performance. 

In comparison to the model proposed in \cite{asc}, which achieved 97.97\%  accuracy on the test set using a CNN model, our baseline model, trained without synthetic data, attained 99.35\% accuracy. This underscores the superior performance of our model architecture over a CNN model. Moreover, the inclusion of extra data through a text-to-speech system resulted in our best-performing model, achieving 99.59\% accuracy. The cited model achieved 97.97\% accuracy on the test set, our top-performing model achieved 99.59\%, resulting in 79.8\% relative reduction and 1.6\% absolute reduction in error rate.

\section{Conclusion and future work}
\label{sec:conc}
This work presented AraSpot for Arabic Spoken keyword Spotting that achieved State-of-the-Art SOTA 99.59\% result outperforming previous approaches, by employing synthetic data generation using text-to-speech, online data augmentation, and introducing ConformerGRU model architecture.
For future work, we recommend expanding the number of commands and increasing the number of speakers to expand the synthetic data generated.

\bibliographystyle{ACM-Reference-Format}
\bibliography{ASR}


\end{document}